\pdfoutput=1

\documentclass[11pt]{article}

\usepackage[final]{acl}
\usepackage{times}
\usepackage{latexsym}

\usepackage{amsmath}
\usepackage{amssymb}
\usepackage{bm}  

\usepackage{graphicx} 
\usepackage{booktabs} 
\usepackage[normalem]{ulem} 
\usepackage[ruled,vlined]{algorithm2e}
\usepackage{rotating}
\usepackage{multirow}
\usepackage{booktabs}
\usepackage{tabularx} 
\usepackage{float}
\usepackage{hyperref}
\hypersetup{colorlinks=true, urlcolor=blue}

\usepackage[T1]{fontenc}

\usepackage[utf8]{inputenc}

\usepackage{microtype}

\usepackage{inconsolata}

\usepackage{graphicx}
\usepackage{caption}  

\title{SLiNT: Structure-aware Language Model with Injection and Contrastive Training for Knowledge Graph Completion}

\author{
Mengxue Yang$^{1}$,
Chun Yang$^{1}$,
Jiaqi Zhu$^{1,2,*}$,
Jiafan Li$^{1,2}$\\
\textbf{Jingqi Zhang}$^{1}$,
\textbf{Yuyang Li}$^{1,3}$,
\textbf{Ying Li}$^{1}$\thanks{*Corresponding author.} \\
$^{1}$University of Chinese Academy of Sciences, Beijing, China \\
$^{2}$Institute of Software, Chinese Academy of Sciences, Beijing, China \\
$^{3}$National Astronomical Observatories, Chinese Academy of Sciences, Beijing, China\\
\texttt{\{yangmengxue20@mails.ucas.ac.cn, zhujq@ios.ac.cn, liying21@ucas.ac.cn\}}
}

\begin{document}
\maketitle
\title{SLiNT: Structure-aware Language Model with Injection and Contrastive Training for Knowledge Graph Completion}

\begin{abstract}
Link prediction in knowledge graphs (KGs) requires integrating structural information and semantic context to infer missing entities. While large language models (LLMs) offer strong generative reasoning capabilities, their limited exploitation of structural signals often results in \emph{structural sparsity} and \emph{semantic ambiguity}, especially under incomplete or zero-shot settings. To address these challenges, we propose \textbf{SLiNT} (\textbf{S}tructure-aware \textbf{L}anguage model with \textbf{I}njection and co\textbf{N}trastive \textbf{T}raining), a modular framework that injects KG-derived structural context into a frozen LLM backbone with lightweight LoRA-based adaptation for robust link prediction. Specifically, \textbf{Structure-Guided Neighborhood Enhancement (SGNE)} retrieves pseudo-neighbors to enrich sparse entities and mitigate missing context; \textbf{Dynamic Hard Contrastive Learning (DHCL)} introduces fine-grained supervision by interpolating hard positives and negatives to resolve entity-level ambiguity; and \textbf{Gradient-Decoupled Dual Injection (GDDI)} performs token-level structure-aware intervention while preserving the core LLM parameters. Experiments on WN18RR and FB15k-237 show that SLiNT achieves superior or competitive performance compared with both embedding-based and generation-based baselines, demonstrating the effectiveness of structure-aware representation learning for scalable knowledge graph completion.
\end{abstract}

\section{Introduction}
Knowledge graphs (KGs) encode real-world facts as structured triples $(h, r, t)$, where $h$ and $t$ denote the \textit{head} and \textit{tail} entities,  respectively, and $r$ is the relation connecting them. As a backbone for structured knowledge representation, KGs empower a variety of downstream applications such as question answering~\cite{DBLP:conf/acl/SaxenaTT20}, recommendation~\cite{DBLP:conf/kdd/Wang00LC19}, and commonsense reasoning~\cite{DBLP:conf/emnlp/LinCCR19}. However, real-world KGs are often incomplete, which motivates the task of \textit{knowledge graph completion} (KGC), i.e., predicting missing entities or relations.

Traditional KGC methods such as TransE~\cite{DBLP:conf/nips/BordesUGWY13}, DistMult~\cite{DBLP:journals/corr/YangYHGD14a}, and RotatE~\cite{DBLP:conf/iclr/SunDNT19} learn low-dimensional embeddings for entities and relations, and rank candidate triples based on geometric scoring functions. While these models perform well in dense regions of the KG, they often underperform on long-tail entities with sparse local neighborhoods. To mitigate this, some extensions incorporate textual features~\cite{DBLP:journals/tacl/WangGZZLLT21} or graph-aware context~\cite{DBLP:conf/iclr/VashishthSNT20}, but still struggle with generalization and semantic discrimination.

\begin{figure}[t]
    \centering
    \includegraphics[width=1.04\linewidth]{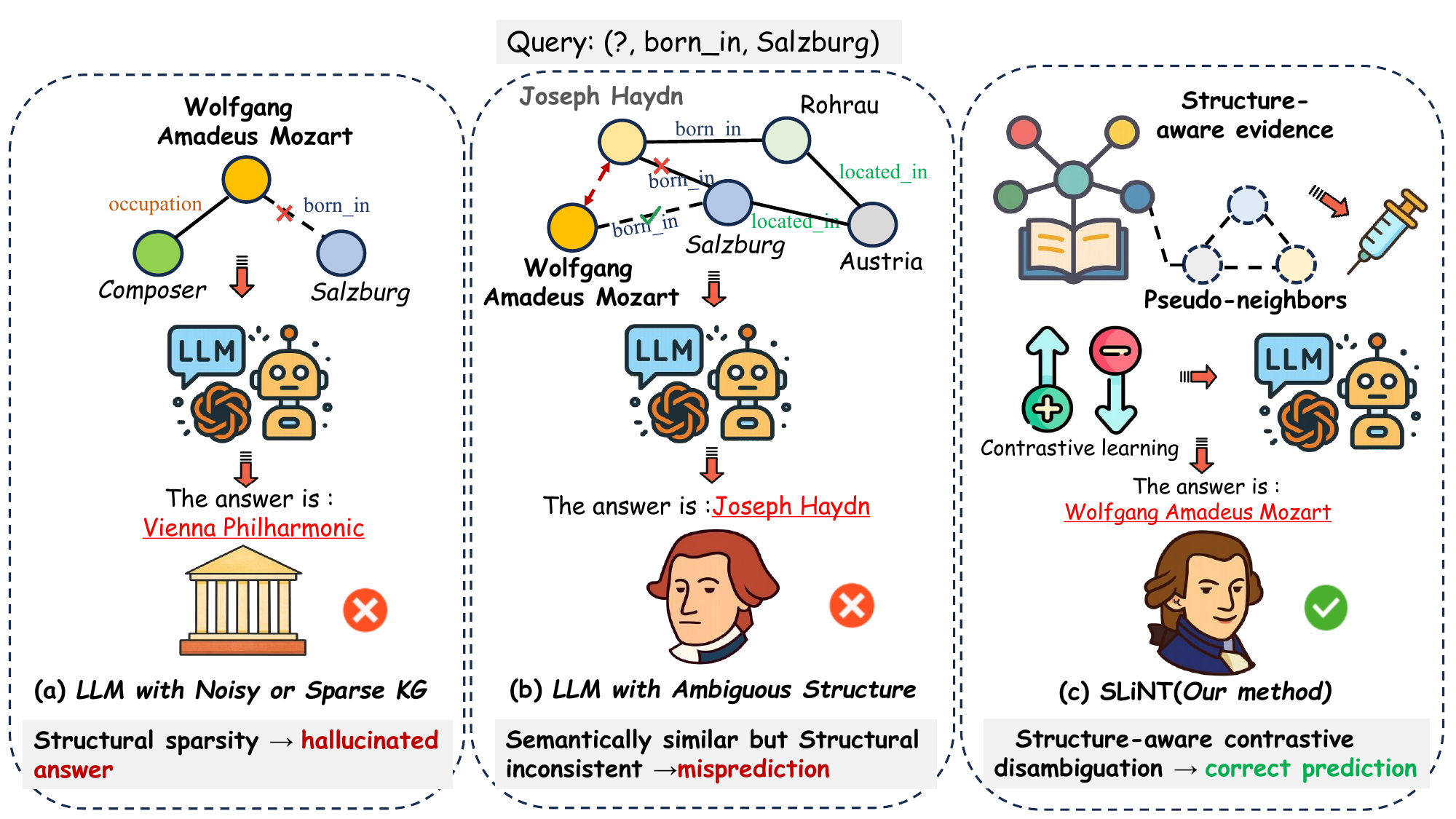}
    \caption{
       Motivating example for \textbf{SLiNT}. Given query \((?, \texttt{born\_in}, \texttt{Salzburg})\), (a) LLMs hallucinate due to sparse KG; (b) Semantic similarity overrides structural correctness, causing misprediction; (c) \textbf{SLiNT} disambiguates candidates via contrastive reasoning and structure injection.
}
    \label{fig:motivation}
\end{figure}

Recent advances in large language models (LLMs) have introduced a new paradigm for knowledge graph completion (KGC), where pretrained models leverage semantic priors to generate missing entities from textualized queries~\cite{DBLP:conf/acl/LewisLGGMLSZ20, DBLP:journals/jmlr/RaffelSRLNMZLL20, DBLP:conf/www/XieZLDCXCC22, DBLP:conf/acl/SaxenaKG22}.
 To improve grounding, several strategies incorporate KG-derived signals into prompts. Instruction tuning~\cite{DBLP:conf/semweb/LiuTSH24} encodes relation semantics and output formats into natural language templates, while structural augmentation~\cite{DBLP:conf/coling/LiuZLYP25, DBLP:journals/corr/abs-2402-02389, DBLP:journals/inffus/YangZMLFZ25} use local subgraphs or structure-aware demonstrations to better align with graph context. Despite these strategies having shown promising improvements in generation controllability and KG-awareness, persistent limitations emerge,  as illustrated in Figure~\ref{fig:motivation}, when examining the link prediction query \((?, \texttt{born\_in}, \texttt{Salzburg})\):
\begin{itemize}
   \item \textbf{Challenge 1: Structural Sparsity} — KG-augmented LLMs rely on local subgraph context for grounding, yet many entities are poorly connected. In this case, sparse links around the gold entity “Wolfgang Amadeus Mozart” offer little structural support, thereby causing the model to hallucinate plausible but unsupported answers such as “Vienna Philharmonic.” This reflects a critical failure: the generation collapses when the KG lacks sufficient structural cues.

    \item \textbf{Challenge 2: Semantic Ambiguity} — Even when structurally valid entities like “Wolfgang Amadeus Mozart” are retrieved, models may make wrong predictions by selecting semantically similar but incorrect alternatives such as “Joseph Haydn.” This confusion arises because current LLMs favor surface-level similarity over structural alignment, lacking mechanisms to resolve fine-grained, relation-specific conflicts in entity semantics.
\end{itemize}

To tackle the aforementioned challenges, we propose \textbf{SLiNT} (\textbf{S}tructure-aware \textbf{L}anguage model with \textbf{I}njection and co\textbf{N}trastive \textbf{T}raining), a unified generative framework that explicitly integrates structural context and fine-grained supervision into a frozen LLM backbone with lightweight LoRA-based adaptation~\cite{DBLP:conf/iclr/HuSWALWWC22}. To address Challenge 1, SLiNT introduces \textbf{Structure-Guided Neighborhood Enhancement (SGNE)}, which retrieves top-$k_s$ pseudo-neighbors from pretrained KG embeddings and fuses them with attention to construct richer contextual representations for sparsely connected entities. To mitigate Challenge 2, we develop \textbf{Dynamic Hard Contrastive Learning (DHCL)}, which synthesizes interpolated hard positives and negatives based on semantic proximity and structural signals, encouraging the model to distinguish structurally coherent answers from semantically similar but misleading distractors. To bridge the gap between structural representations and language generation, we further design \textbf{Gradient-Decoupled Dual Injection (GDDI)}, which injects the enhanced structural representations into the frozen LLM backbone at the token level through prompt-based augmentation and substitution, while leveraging lightweight LoRA adaptation to maintain parameter efficiency. Together, these components enable SLiNT to perform robust link prediction under both sparse and ambiguous KG scenarios, while maintaining generation fluency, structural faithfulness, and parameter efficiency. Our main contributions are summarized as follows:

\begin{itemize}
    \item We propose \textbf{SLiNT}, the first structure-aware generative framework that jointly integrates pseudo-neighbor enhancement, contrastive disambiguation, and token-level structure injection into a frozen LLM backbone for link prediction.
    \item We introduce two novel techniques to realize the framework: \textbf{DHCL}, for structure-aware contrastive learning, and \textbf{GDDI}, a lightweight gradient-decoupled injection mechanism.
    \item We empirically show that SLiNT achieves superior or competitive performance on two standard benchmarks, while maintaining robustness in sparse and ambiguous KG scenarios.
\end{itemize}

\section{Related Work}
\label{sec:related}
Prior work on knowledge graph completion (KGC) can be broadly categorized into two paradigms: embedding-based models and generation-based models.

\paragraph{Embedding-based KGC.} 
Classical models such as TransE~\cite{DBLP:conf/nips/BordesUGWY13}, DistMult~\cite{DBLP:journals/corr/YangYHGD14a}, and RotatE~\cite{DBLP:conf/iclr/SunDNT19} encode entities and relations into continuous vector spaces, scoring triples based on distance or semantic compatibility. While efficient and interpretable, these models depend heavily on dense local structures and struggle with long-tail or sparsely connected entities. Later extensions incorporate auxiliary textual~\cite{DBLP:journals/tacl/WangGZZLLT21} or structural~\cite{DBLP:conf/iclr/VashishthSNT20} information to improve robustness, but remain limited in handling diverse or ambiguous semantics.

\paragraph{Generation-based KG Completion.}
Unlike embedding-based methods that learn entity and relation vectors, generation-based approaches formulate KG completion as a text generation task. Early works such as KGT5~\cite{DBLP:conf/acl/SaxenaKG22}, GenKGC~\cite{DBLP:conf/www/XieZLDCXCC22} and KG-S2S~\cite{DBLP:conf/coling/ChenWLL22} recast triple prediction into sequence-to-sequence learning, enabling flexible reasoning over natural language. Later models, including GS-KGC~\cite{DBLP:journals/inffus/YangZMLFZ25} and FtG~\cite{DBLP:conf/coling/LiuZLYP25}, incorporate structural features from subgraphs to provide auxiliary context, while KICGPT~\cite{DBLP:journals/corr/abs-2402-02389} and DIFT~\cite{DBLP:conf/semweb/LiuTSH24} adopt instruction tuning and in-context demonstrations to improve generation quality. These approaches significantly advance LLM-based KGC by leveraging pretrained language priors. However, they typically lack mechanisms to model structural ambiguity or decision boundaries, limiting their performances in sparse or confusing regions.

\section{Methodology}
\begin{figure*}[t]
\centering
\includegraphics[width=1.02\linewidth]{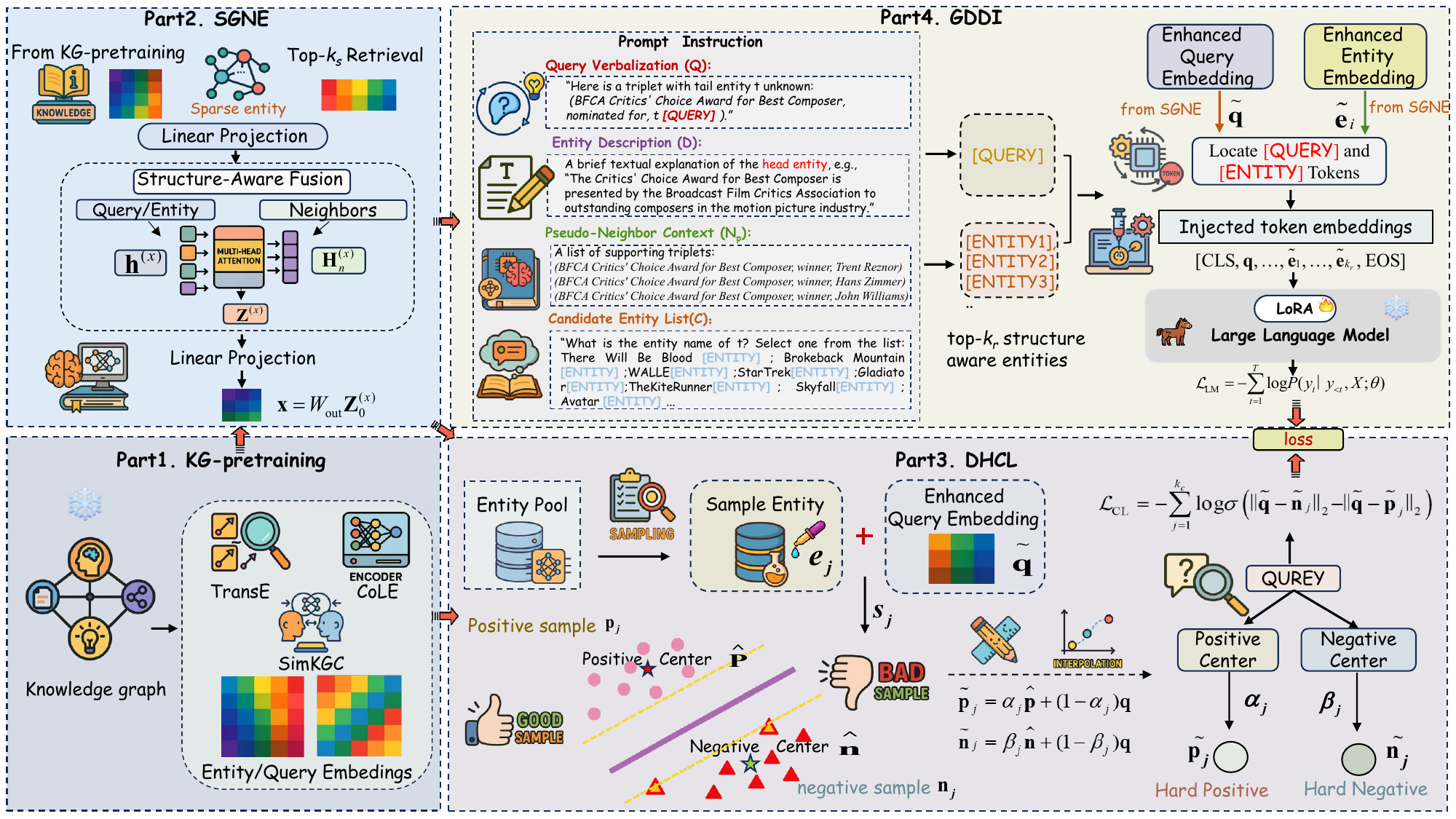}
\caption{
\textbf{Overview of the SLiNT framework.} 
It consists of \textbf{SGNE} for neighbor-enhanced fusion, 
\textbf{DHCL} for contrastive augmentation, 
and \textbf{GDDI} for structural injection with LoRA fine-tuning.
}
\label{fig:framework}
\end{figure*}

\textbf{SLiNT} tackles two key challenges in knowledge graph completion (KGC)—\textit{structural sparsity} and \textit{semantic ambiguity}—via three modules: structure-guided neighborhood enhancement, contrastive learning, and structure-aware injection. As shown in Figure~\ref{fig:framework}, the pipeline starts from pretrained KG embeddings, sequentially applies enhancement and contrastive supervision, then injects structure-derived signals into a frozen LLM backbone, while fine-tuning only the lightweight LoRA adapters for efficient adaptation.

\paragraph{Problem Formulation.} 
We formalize the task as link prediction over a knowledge graph \( \mathcal{G} = (\mathcal{E}, \mathcal{R}, \mathcal{T}) \), where \( \mathcal{T} \subseteq \mathcal{E} \times \mathcal{R} \times \mathcal{E} \) denotes a set of factual triples. 
Given an incomplete query \( q = (h, r, ?) \) or \( (?, r, t) \), the objective is to find the missing entity, either the head \(h\) or the tail \(t\), from the entity set \(\mathcal{E}\). Following DIFT, we adopt a two-stage formulation. A pretrained KG embedding model \( M_E \) is first used to rank all candidate entities, producing a top-$m$ list:
\begin{equation}
\mathcal{C}(q) = \operatorname{Top\text{-}m}(M_E(q)) = [e_1, e_2, \dots, e_m],
\label{eq:candidate_selection}
\end{equation}
where each candidate \( e_i \) is associated with an embedding \( \mathbf{e}_i \in \mathbb{R}^d \), and the query is represented by \( \mathbf{q} \in \mathbb{R}^d \). These structural representations are used as inputs to the subsequent SLiNT modules.

\subsection{Structure-Guided Neighborhood Enhancement (SGNE)}
To address structural sparsity, SGNE enhances each input embedding, either a query \( \mathbf{q} \in \mathbb{R}^d \) or a candidate entity \( \mathbf{e}_i \in \mathbb{R}^d \), by aggregating its top-\(k_s\) \textit{structural pseudo-neighbors}, i.e., nearest neighbors in the pretrained KG embedding space rather than true KG neighbors. We retrieve these pseudo-neighbors from a global entity pool \( \mathcal{E} \in \mathbb{R}^{N \times d} \) based on cosine similarity:
\begin{equation}
\mathcal{N}_p(\mathbf{x}) = \text{Top-}k_s\left( \cos(\mathbf{x}, \mathcal{E}) \right), \quad \mathbf{x} \in \{ \mathbf{q}, \mathbf{e}_i \}.
\label{eq:sgne_topk}
\end{equation}

Let \( \mathbf{E}_n^{(x)} \in \mathbb{R}^{k_s \times d} \) denote the corresponding embedding matrix of pseudo-neighbors. We project both the input and its neighbors into a shared latent space using a learnable matrix \( W_{\text{in}} \in \mathbb{R}^{d \times h} \), followed by a SiLU activation $\phi$ ~\cite{DBLP:journals/nn/ElfwingUD18}:
\begin{equation}
\mathbf{h}^{(x)} = \phi(W_{\text{in}} \mathbf{x}), \quad 
\mathbf{H}_n^{(x)} = \phi(W_{\text{in}} \mathbf{E}_n^{(x)}).
\label{eq:sgne_projection}
\end{equation}

We concatenate the input and neighbor representations, apply the multi-head attention:
\begin{equation}
\mathbf{Z}^{(x)} = \operatorname{MultiHeadAttn} \left( \mathbf{h}^{(x)} \, \Vert \, \mathbf{H}_n^{(x)} \right).
\label{eq:sgne_attn}
\end{equation}

The enhanced representation is obtained by extracting the first token and projecting it back to the output dimension via \( W_{\text{out}} \in \mathbb{R}^{h \times d'} \):
\begin{equation}
\tilde{\mathbf{x}} = W_{\text{out}} \mathbf{Z}_0^{(x)}, \quad \tilde{\mathbf{x}} \in \{\tilde{\mathbf{q}}, \tilde{\mathbf{e}_i} \}.
\label{eq:sgne_output}
\end{equation}

The final outputs \( \tilde{\mathbf{q}}, \tilde{\mathbf{e}}_i \in \mathbb{R}^{d'} \) are used for contrastive learning and token-level generation in subsequent modules.

\subsection{Dynamic Hard Contrastive Learning (DHCL)}
\label{sec:dhcl}
While SGNE enhances structural representations, it lacks supervision to distinguish structurally coherent entities from semantically similar but structurally divergent distractors. To address this, we propose Dynamic Hard Contrastive Learning (DHCL), a structure-sensitive contrastive objective that promotes fine-grained discrimination in the structural space. Rather than comparing raw entity pairs, DHCL interpolates between the query and structure-derived prototypes to generate boundary-level hard positives and negatives, simulating ambiguous decisions near structural margins.
This encourages the model to favor structurally valid answers and suppress misleading semantic lookalikes.
The full procedure is shown in Algorithm~\ref{alg:dhcl}.

\begin{algorithm}[t]
\caption{Dynamic Hard Contrastive Learning (DHCL)}
\label{alg:dhcl}
\KwIn{
    Enhanced query $\tilde{\mathbf{q}} \in \mathbb{R}^{d'}$; \\
    Entity pool $\mathcal{E} \in \mathbb{R}^{N \times d'}$; \\
    Sample size $N_c$; \\
    Contrastive sample size $k_c$
}
\KwOut{Contrastive loss $\mathcal{L}_{\text{CL}}$}

\BlankLine
Sample $N_c$ entities $\{\mathbf{e}_j\}_{j=1}^{N_c} \subset \mathcal{E}$\;
Compute cosine similarity $s_j \leftarrow \cos(\tilde{\mathbf{q}}, \mathbf{e}_j)$\;
Select positives $\mathcal{P}_c \leftarrow \operatorname{Top\text{-}k_c}^{\text{high}}(\{s_j\})$\;
Select negatives $\mathcal{N}_c \leftarrow \operatorname{Top\text{-}k_c}^{\text{low}}(\{s_j\})$\;
Compute prototype centers $\hat{\mathbf{p}}, \hat{\mathbf{n}}$\;
Interpolate hard samples $\tilde{\mathbf{p}}_j, \tilde{\mathbf{n}}_j$\;
Compute contrastive loss $\mathcal{L}_{\text{CL}}$\;
\Return $\mathcal{L}_{\text{CL}}$\;
\end{algorithm}

Given a query embedding \( \mathbf{q} \), enhanced via SGNE as \( \tilde{\mathbf{q}} \), we randomly sample \( N_c \) entities \( \{\mathbf{e}_j\} \) from the global entity pool \( \mathcal{E} \), and compute their cosine similarities:
\begin{equation}
s_j = \cos(\tilde{\mathbf{q}}, \mathbf{e}_j) = \frac{\tilde{\mathbf{q}}^\top \mathbf{e}_j}{\|\tilde{\mathbf{q}}\| \cdot \|\mathbf{e}_j\|}.
\end{equation}

\paragraph{Hard Sample Mining.}
To provide contrastive supervision, the top-$k_c$ most similar and least similar entities are selected:
\begin{align}
\mathcal{P}_c &= \operatorname{Top\text{-}\mathit{k}_c}^{\text{high}}(\{s_j\}), \quad \mathcal{N}_c = \operatorname{Top\text{-}\mathit{k}_c}^{\text{low}}(\{s_j\}),
\end{align}
where $\mathcal{P}_c = \{\mathbf{p}_j\}$ and $\mathcal{N}_c = \{\mathbf{n}_j\}$ denote the contrastive positive and negative entity sets, respectively. The hyperparameter $k_c$ specifies the number of hard samples used for contrastive training.

\paragraph{Prototype Construction.}
To obtain representative prototypes for contrastive learning, we compute weighted centers:
\begin{equation}
\hat{\mathbf{p}} = \frac{\sum_j w_j^+ \mathbf{p}_j}{\sum_j w_j^+}, \quad \hat{\mathbf{n}} = \frac{\sum_j w_j^- \mathbf{n}_j}{\sum_j w_j^-},
\end{equation}
where \( w_j^+, w_j^- \in \mathbb{R}^+ \) are sampled from a uniform distribution.

\paragraph{Interpolation.}
To generate boundary-sensitive examples, interpolation is performed between the  query and the prototype centers:
\begin{equation}
\begin{aligned}
\tilde{\mathbf{p}}_j &= \alpha_j \hat{\mathbf{p}} + (1 - \alpha_j) \mathbf{q}, \\
\tilde{\mathbf{n}}_j &= \beta_j \hat{\mathbf{n}} + (1 - \beta_j) \mathbf{q},
\end{aligned}
\end{equation}
where \( \alpha_j, \beta_j \in [0, 1] \) are interpolation coefficients sampled uniformly. We constrain the coefficients to $\alpha_j, \beta_j \sim \mathcal{U}(0.3, 0.7)$ in practice. These synthetic samples approximate near-boundary contrasts in the structure space.

\paragraph{Contrastive Loss.}
We then optimize a margin-based contrastive loss over interpolated examples:
\begin{equation}
\mathcal{L}_{\text{CL}} = - \sum_{j=1}^{k_c} 
\log \sigma \left( \| \tilde{\mathbf{q}} - \tilde{\mathbf{n}}_j \|_2 - \| \tilde{\mathbf{q}} - \tilde{\mathbf{p}}_j \|_2 \right).
\end{equation}

\vspace{1mm}
\noindent
This encourages the model to align query representations with structurally consistent entities while pushing away structurally incompatible ones, improving fine-grained discrimination in structure-aware settings.


\subsection{Gradient-Decoupled Dual Injection (GDDI)}
We propose Gradient-Decoupled Dual Injection (GDDI), a dual mechanism that incorporates structure-enhanced representations into a frozen LLM backbone through \textit{prompt-level augmentation} and \textit{token-level injection}, while fine-tuning only the lightweight LoRA adapters. This design allows the model to leverage KG-derived context efficiently, without updating the full set of LLM parameters.

\paragraph{Prompt Construction.}
For each query triple \( q = (h, r, ?) \) or \( (?, r, t) \), we construct a generation prompt \( \mathcal{P}(q) \) by concatenation:
\begin{equation}
\mathcal{P}(q) = [\mathcal{Q}; \mathcal{D}; \mathcal{N}_p; \mathcal{C}],
\end{equation}
Where \( \mathcal{Q} \) is a natural language verbalization of the query (e.g., ``(BFCA Critics' Choice Award for Best Composer, nominated for, ?)''), \( \mathcal{D} \) provides a brief textual description of the known entity (either head or tail), \( \mathcal{N}_p \) contains pseudo-neighbor triplets retrieved by SGNE, which are constructed by mapping structural pseudo-neighbors (retrieved in the embedding space) back to KG triples, and \( \mathcal{C} \) lists the top-\( m \) ranked candidates from the KG embedding model.
An illustrative example from the FB15k-237 dataset is provided in Figure~\ref{fig:framework}.

\paragraph{Token-Level Injection.}
After constructing the prompt, we identify the token positions of the \texttt{[QUERY]} and top-\(k_r\) \texttt{[ENTITY]} markers. We inject SGNE-enhanced embeddings at these locations:
\begin{equation}
\mathbf{E}_{\text{input}}[p_{\text{query}}] = \tilde{\mathbf{q}}, \quad
\mathbf{E}_{\text{input}}[p_{\text{entity}}^{(i)}] = \tilde{\mathbf{e}}_i,
\end{equation}
where \( \tilde{\mathbf{q}}, \tilde{\mathbf{e}}_i \in \mathbb{R}^{d'} \) are structure-aware embeddings, and \( p_{\text{query}}, p_{\text{entity}}^{(i)} \) denote the corresponding token indices.

\paragraph{Training.}
We adapt the frozen LLM  backbone via parameter-efficient LoRA~\cite{DBLP:conf/iclr/HuSWALWWC22}, where only LoRA adapters are trainable. This dual mechanism strengthens the model’s capacity to integrate structure-aware representations during generation, enabling more accurate entity disambiguation in sparse and ambiguous contexts.
\subsection{Training Objective}

The training objective combines language modeling with structure-aware contrastive supervision:

\begin{equation}
\mathcal{L}_{\text{total}} = \mathcal{L}_{\text{LM}} + \lambda \cdot \mathcal{L}_{\text{CL}},
\end{equation}
where \( \lambda \in \mathbb{R}^+ \) balances the generation loss \( \mathcal{L}_{\text{LM}} \) and the contrastive loss \( \mathcal{L}_{\text{CL}} \). The language modeling loss is defined as:

\begin{equation}
\mathcal{L}_{\text{LM}} = - \sum_{t=1}^T \log P(y_t \mid y_{<t}, X; \theta),
\end{equation}
where \( y_t \) is the target token at timestep \( t \), \( y_{<t} \) is the partial output sequence, \( X \) is the structure-injected input, and \( \theta \) denotes the LLM parameters. This ensures that the generation is conditioned not only on textual prompts but also on injected structural signals.
Detailed theoretical analyses of retrieval complexity, the stability of the contrastive loss, and structural injection alignment are provided in Appendix~\ref{appendix:complexity}, \ref{appendix:contrastive}, and \ref{appendix:alignment}.

\section{Experiments}
We conduct comprehensive experiments to evaluate the effectiveness of \textbf{SLiNT} on two widely used knowledge graph completion (KGC) benchmarks: \textbf{FB15k-237}~\cite{DBLP:conf/emnlp/ToutanovaCPPCG15} and \textbf{WN18RR}~\cite{DBLP:conf/aaai/DettmersMS018}. The statistics analysis of these datasets is provided in Appendix~\ref{appendix:dataset-sparsity}. Our experiments aim to answer the following research questions:

\begin{itemize}
    \item \textbf{RQ1}: Does SLiNT outperform state-of-the-art embedding-based and generation-based KGC methods?
    \item \textbf{RQ2}: What are the individual contributions of SGNE, DHCL, and GDDI to the overall performance?
    \item \textbf{RQ3}: How robust is SLiNT under low-resource or structurally incomplete KG scenarios?
\end{itemize}

\subsection{Experimental Setup}

\paragraph{Baselines.} We compare \textbf{SLiNT} with two categories of methods: (1) embedding-based models such as TransE~\cite{DBLP:conf/nips/BordesUGWY13}, RotatE~\cite{DBLP:conf/iclr/SunDNT19}, and others; and (2) generation-based models including DIFT~\cite{DBLP:conf/semweb/LiuTSH24}, KICGPT~\cite{DBLP:journals/corr/abs-2402-02389}, and others. A full list of baselines is provided in Table~\ref{tab:main_results_final}.

\paragraph{Backbone Configurations.}
 We instantiate SLiNT with three representative structural backbones:
(1) \textbf{SLiNT + TransE} uses a translational embedding model~\cite{DBLP:conf/nips/BordesUGWY13} that captures relational regularities through distance-based scoring, representing a lightweight and widely-used baseline for structure-only supervision;
(2) \textbf{SLiNT + SimKGC} builds on contrastive representation learning~\cite{DBLP:conf/acl/0046ZWL22}, enhancing entity discrimination by aligning positive pairs and pushing apart hard negatives;
(3) \textbf{SLiNT + CoLE} leverages a hybrid encoder~\cite{DBLP:conf/cikm/LiuSLH22} combining local neighborhood information with pre-trained textual features, providing rich structure-semantic fusion.

\paragraph{Evaluation Metrics.} 
We adopt standard evaluation metrics commonly used in knowledge graph completion tasks, including \textbf{Mean Reciprocal Rank (MRR)} and \textbf{Hits@K} (with $K = 1, 3, 10$). 
\textbf{MRR} measures the average inverse rank of the correct entity across all test queries, providing a fine-grained assessment of ranking performance. 
\textbf{Hits@K} reports the proportion of test queries for which the correct entity appears within the top-$K$ ranked candidates, reflecting the model's ability to retrieve relevant entities. 
Higher MRR and Hits@K scores indicate better predictive accuracy. 
These metrics are computed under the standard filtered setting, where corrupted triples that already exist in the KG are excluded during ranking to ensure fair evaluation.

\subsection{Implementation Details}
We implement SLiNT using PyTorch with mixed-precision training on 8×64GB MetaX GPUs (performance comparable to A100s). All experiments leverage frozen LLaMA-7B~\footnote{\url{https://huggingface.co/meta-llama/Llama-2-7b-chat-hf}} with pre-trained KG embeddings (TransE, SimKGC, CoLE), and employ LoRA for lightweight adaptation. Key training hyperparameters include a batch size of 64, a learning rate of $2 \times 10^{-5}$, and contrastive loss weight $\lambda=0.5$. Further configuration details are provided in Appendix~\ref{appendix:training}.





\subsection{Main Results (RQ1)}
We evaluate SLiNT on FB15k-237 and WN18RR, comparing it against two major categories of methods: \textit{embedding-based} models and \textit{generation-based} models. Results are reported in Table~\ref{tab:main_results_final}.

\paragraph{Overall Performance.} 
SLiNT achieves superior or competitive performance across both datasets. 
On \textbf{FB15k-237}, \textbf{SLiNT + CoLE} attains the highest MRR (0.443) and strong Hits@1 (0.368), while slightly trailing NBFNet in Hits@10 (0.599 vs.\ 0.591). 
On \textbf{WN18RR}, \textbf{SLiNT + SimKGC} yields the best MRR (0.691) and Hits@1 (0.626), with Hits@10 of 0.805, outperforming other generation-based baselines. 
Although NCRL achieves a higher Hits@10 (0.850), SLiNT + SimKGC maintains the best overall balance with superior MRR. 
Compared with vanilla LLaMA variants, SLiNT delivers notable MRR gains: +0.205 on FB15k-237 and +0.252 on WN18RR, highlighting the benefits of structure-aware injection and contrastive learning.
\paragraph{Comparison with Prior Methods.} Embedding-based methods such as NBFNet and SimKGC perform well on WN18RR but underperform on long-tail relations on FB15k-237. Generation-based baselines like DIFT and KICGPT improve semantic controllability, but their reliance on template-based augmentation limits their robustness. SLiNT outperforms all prior generative models under identical KG embeddings (e.g., CoLE), highlighting its superior capacity to capture structure-aware semantics.

\paragraph{SLiNT Variants.} 
SLiNT yields consistent improvements across all KG embeddings, TransE, SimKGC, and CoLE, validating its robustness and \textit{plug-and-play} compatibility with frozen LLM backbone. Rather than relying on any specific encoder, SLiNT adapts flexibly to different structural priors. Case studies in Appendix~\ref{appendix:case} further illustrate how it resolves fine-grained ambiguities through structure-aware supervision.

\begin{table*}[ht]
\centering
\resizebox{\textwidth}{!}{
\begin{tabular}{lcccccccccc}
\toprule
\multirow{2}{*}{\textbf{Models}}  & \multicolumn{4}{c}{\textbf{FB15K-237}} & \multicolumn{4}{c}{\textbf{WN18RR}} \\
\cmidrule(lr){2-5} \cmidrule(lr){6-9}
& \textbf{MRR} & \textbf{Hits@1} & \textbf{Hits@3} & \textbf{Hits@10} & \textbf{MRR} & \textbf{Hits@1} & \textbf{Hits@3} & \textbf{Hits@10} \\
\midrule

\multicolumn{9}{c}{\textbf{Embedding-based}} \\
\midrule
TransE~\cite{DBLP:conf/nips/BordesUGWY13} & 0.312 & 0.212 & 0.354 & 0.510 & 0.225 & 0.016 & 0.403 & 0.521 \\
RotatE~\cite{DBLP:conf/iclr/SunDNT19} & 0.338 & 0.241 & 0.375 & 0.533 & 0.476 & 0.428 & 0.492 & 0.571 \\
TuckER~\cite{DBLP:conf/emnlp/BalazevicAH19} & 0.358 & 0.266 & 0.394 & 0.544 & 0.470 & 0.443 & 0.482 & 0.526 \\
Neural-LP~\cite{DBLP:conf/nips/YangYC17} & 0.237 & 0.173 & 0.259 & 0.361 & 0.381 & 0.368 & 0.386 & 0.408 \\
NCRL~\cite{DBLP:conf/iclr/ChengAS23} & 0.300 & --- & --- & 0.473 & 0.670 & 0.563 & --- & \textbf{0.850} \\
CompGCN~\cite{DBLP:conf/iclr/VashishthSNT20} & 0.355 & 0.264 & 0.390 & 0.535 & 0.479 & 0.443 & 0.494 & 0.546 \\
HittER~\cite{DBLP:conf/emnlp/ChenLG0ZJ21} & 0.373 & 0.279 & 0.409 & 0.558 & 0.503 & 0.462 & 0.516 & 0.584 \\
NBFNet~\cite{DBLP:conf/nips/ZhuZXT21} & 0.415 & 0.321 & 0.454&\textbf{0.599} & 0.551 & 0.497 & 0.573 & 0.666 \\

\midrule
KG-BERT~\cite{DBLP:journals/corr/abs-1909-03193} & --- & --- & --- & 0.420 & 0.216 & 0.041 & 0.302 & 0.524 \\
StAR~\cite{DBLP:conf/www/WangSLZW021} & 0.365 & 0.266 & 0.404 & 0.562 & 0.551 & 0.459 & 0.594 & 0.732 \\
MEM-KGC~\cite{DBLP:journals/access/ChoiJK21} & 0.346 & 0.253 & 0.381 & 0.531 & 0.557 & 0.475 & 0.604 & 0.704 \\
SimKGC~\cite{DBLP:conf/acl/0046ZWL22} & 0.338 & 0.252 & 0.364 & 0.511 & 0.671& 0.595 & 0.719& 0.802\\
CoLE~\cite{DBLP:conf/cikm/LiuSLH22} & 0.389 & 0.294 & 0.429 & 0.572 & 0.593 & 0.538 & 0.616 & 0.701 \\

\midrule
\multicolumn{9}{c}{\textbf{Generation-based}} \\
\midrule
GenKGC\cite{DBLP:conf/www/XieZLDCXCC22} & --- & 0.192 & 0.355 & 0.439 & --- & 0.287 & 0.403 & 0.535 \\
KGT5~\cite{DBLP:conf/acl/SaxenaKG22} & 0.276 & 0.210 & --- & 0.414 & 0.508 & 0.487 & --- & 0.544 \\
KG-S2S~\cite{DBLP:conf/coling/ChenWLL22} & 0.336 & 0.257 & 0.373 & 0.498 & 0.574 & 0.531 & 0.595 & 0.661 \\
ChatGPT$_{\text{one-shot}}$~\cite{openai2023chatgpt} & --- & 0.267 & --- & --- & --- & 0.212 & --- & --- \\
KICGPT~\cite{DBLP:journals/corr/abs-2402-02389} & 0.412 &0.327 &0.448& 0.581 & 0.564 & 0.478 & 0.612 & 0.677 \\
LLaMA + TransE~\cite{DBLP:conf/semweb/LiuTSH24} & 0.232 & 0.080 & 0.321 & 0.502 & 0.202 & 0.037 & 0.360 & 0.516 \\
LLaMA + SimKGC~\cite{DBLP:conf/semweb/LiuTSH24} & 0.236 & 0.074 & 0.335 & 0.503 & 0.391 & 0.065 & 0.695 & 0.798 \\
LLaMA + CoLE~\cite{DBLP:conf/semweb/LiuTSH24} & 0.238 & 0.087 & 0.387 & 0.561 & 0.374 & 0.117 & 0.602 & 0.697 \\
DIFT + TransE~\cite{DBLP:conf/semweb/LiuTSH24} & 0.389 & 0.322 & 0.408 & 0.525 & 0.491 & 0.462 & 0.496 & 0.560 \\
DIFT + SimKGC~\cite{DBLP:conf/semweb/LiuTSH24} & 0.402 & 0.338 & 0.418 & 0.528 & \underline{0.686} & \underline{0.616} & \underline{0.730} & \underline{0.806}\\
DIFT + CoLE~\cite{DBLP:conf/semweb/LiuTSH24} &\underline{0.439}& \underline{0.364}& \underline{0.468} & 0.586 & 0.617 & 0.569 &0.638& 0.708 \\

\midrule
\textbf{SLiNT + TransE} & 0.395 & 0.329 & 0.416 & 0.522 & 0.506 & 0.482 & 0.508 & 0.567 \\
\textbf{SLiNT + SimKGC} & 0.416 & 0.355 & 0.433 & 0.529 & \textbf{0.691} & \textbf{0.626} & \textbf{0.731} & 0.805\\
\textbf{SLiNT + CoLE} & \textbf{0.443} & \textbf{0.368} & \textbf{0.472} & \underline{0.591} &0.626 & 0.578 & 0.646 & 0.718\\
\bottomrule
\end{tabular}
}
\caption{Link prediction results on FB15k-237 and WN18RR. Best results are in \textbf{bold} and second-best are \underline{underlined}. We reproduce the results of TransE, SimKGC, and CoLE using their source codes and hyperparameters. The results of other baselines are obtained from their respective original papers.}
\label{tab:main_results_final}
\end{table*}

\subsection{Ablation Study (RQ2)}
To assess the contribution of each component in SLiNT, we conduct ablation experiments on FB15k-237 and WN18RR using CoLE embeddings. Table~\ref{tab:ablation_results} reports the results when removing SGNE, DHCL, or GDDI.

On FB15k-237, removing SGNE leads to a noticeable drop in MRR (0.443 → 0.429), highlighting the value of pseudo-neighbor fusion for structural enrichment. The absence of DHCL causes the largest decline in Hits@1 (0.368 → 0.329), underscoring its role in differentiating close structural candidates and reinforcing fine-grained decision boundaries through structure-aware contrastive training. Removing GDDI also degrades performance, albeit moderately, indicating that token-level structure injection provides complementary gains. Similar results appear on WN18RR, where disabling DHCL again leads to the greatest drop in Hits@1 (0.578 → 0.546), confirming its central role in optimizing entity-level decision boundaries. The consistent declines when omitting SGNE or GDDI further support the necessity of all the three components. Overall, these results demonstrate that each module contributes uniquely to SLiNT’s effectiveness, and their integration is essential for accurate and structure-aware link prediction in challenging KG scenarios.

\begin{table}[t]
\centering
\scriptsize
\renewcommand{\arraystretch}{0.95}
\setlength{\tabcolsep}{3pt}
\begin{tabular}{c cccc|cccc}
\toprule
\textbf{Config} 
& \multicolumn{4}{c|}{\textbf{FB15k-237}} 
& \multicolumn{4}{c}{\textbf{WN18RR}} \\
& MRR & H@1 & H@3 & H@10 & MRR & H@1 & H@3 & H@10 \\
\midrule
Full & \textbf{0.443} & \textbf{0.368} & \textbf{0.472} & \textbf{0.591} & \textbf{0.626} & \textbf{0.578} & \textbf{0.646} & \textbf{0.718} \\
w/o SGNE & 0.429 & 0.342 & 0.453 & 0.572 & 0.612 & 0.560 & 0.630 & 0.707 \\
w/o DHCL & 0.419 & 0.329 & 0.444 & 0.564 & 0.606 & 0.546 & 0.621 & 0.705 \\
w/o GDDI & 0.433 & 0.352 & 0.457 & 0.577 & 0.615 & 0.567 & 0.638 & 0.708 \\
\bottomrule
\end{tabular}
\vspace{-1mm}
\caption{
 Ablation results of \textbf{SLiNT} on FB15k-237 and WN18RR using CoLE. Each module contributes to overall performance.
}
\label{tab:ablation_results}
\vspace{-2mm}
\end{table}

\subsection{Robustness Analysis (RQ3)}
\begin{figure}[h]
\centering
\includegraphics[width=1.05\linewidth]{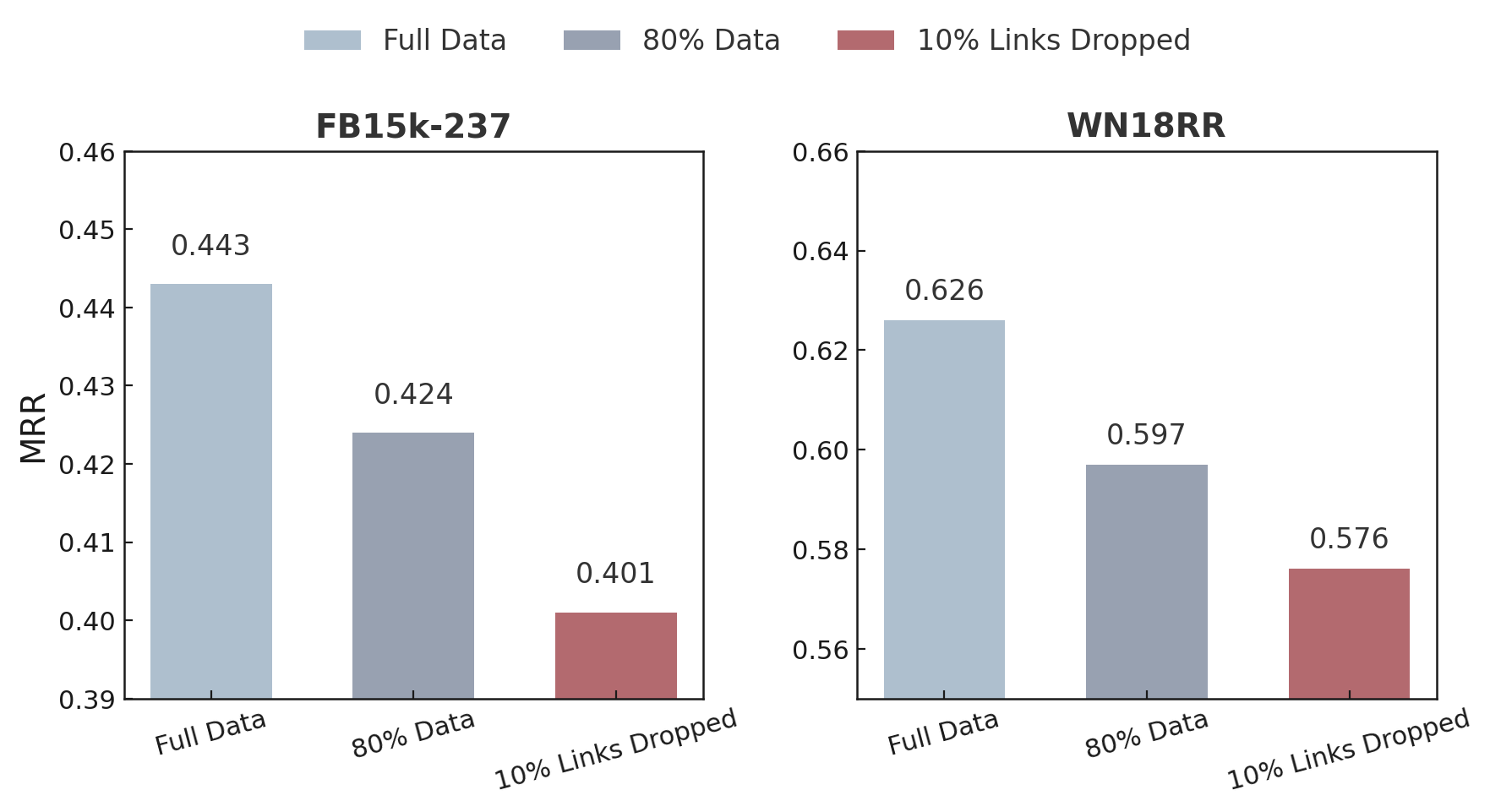}
\caption{Performance of \textbf{SLiNT + CoLE} under limited supervision and structural incompleteness.}
\label{fig:robustness}
\end{figure}

We evaluate SLiNT’s robustness under two common types of knowledge graph sparsity: limited training supervision and incomplete structural connectivity. All experiments are conducted with \textbf{SLiNT+CoLE}. Specifically, we simulate low-resource settings by (1) reducing the training data to 80\%, and (2) randomly removing 10\% of KG edges. As shown in Figure~\ref{fig:robustness}, SLiNT maintains strong MRR under both conditions across FB15k-237 and WN18RR. Performance drops are marginal, confirming its stability against supervision loss and structural noise.

We further assess robustness under degree sparsity. Notice that both datasets exhibit long-tail degree distributions. As shown in Table~\ref{tab:sparsity_stats}, 36.7\% of entities in WN18RR and 22.3\% in FB15k-237 fall within the bottom 20\% of degree, motivating our structure-guided design in SLiNT.

\begin{table}[h]
\centering
\footnotesize
\setlength{\tabcolsep}{3.5pt}
\renewcommand{\arraystretch}{1.0}
\begin{tabular}{lrrrr}
\toprule
\textbf{Dataset} & \textbf{\#Ent.} & \textbf{AvgDeg} & \textbf{LowDeg (\%)} & \textbf{Max/Min} \\
\midrule
WN18RR    & 40,943 & 4.2  & 36.7 & 221 / 1 \\
FB15k-237 & 14,541 & 27.8 & 22.3 & 4,622 / 1 \\
\bottomrule
\end{tabular}
\vspace{-0.4em}
\caption{Structural sparsity statistics. “LowDeg (\%)” denotes the proportion of entities in the bottom 20\% of degree.}
\label{tab:sparsity_stats}
\end{table}

We then evaluate SLiNT on test queries whose gold entities belong to this low-degree group. As shown in Table~\ref {tab:lowdeg-results}, SLiNT consistently outperforms strong baselines across both datasets, highlighting its superior generalization to structurally under-connected entities. These findings support the effectiveness of SGNE and DHCL in real-world incomplete graphs.

\begin{table}[t]
\centering
\scriptsize
\setlength{\tabcolsep}{3pt}
\renewcommand{\arraystretch}{0.95}
\begin{tabular}{c ccc|ccc}
\toprule
\textbf{Model} 
& \multicolumn{3}{c|}{\textbf{FB15k-237}} 
& \multicolumn{3}{c}{\textbf{WN18RR}} \\
& H@1 & H@10 & MRR & H@1 & H@10 & MRR \\
\midrule
LLaMA + CoLE & 0.032 & 0.438 & 0.159 & 0.041 & 0.598 & 0.257 \\
DIFT + CoLE  & 0.281 & 0.497 & 0.379 & 0.502 & 0.646 & 0.571 \\
\textbf{SLiNT + CoLE} & \textbf{0.297} & \textbf{0.511} & \textbf{0.391} & \textbf{0.517} & \textbf{0.658} & \textbf{0.573} \\
\bottomrule
\end{tabular}
\vspace{-1mm}
\caption{
 Performance on low-degree entities (bottom 20\% degree group) from FB15k-237 and WN18RR.
}
\label{tab:lowdeg-results}
\vspace{-2mm}
\end{table}

\subsection{Further Analysis}
\begin{figure}[t]
    \centering
    \includegraphics[width=1\linewidth]{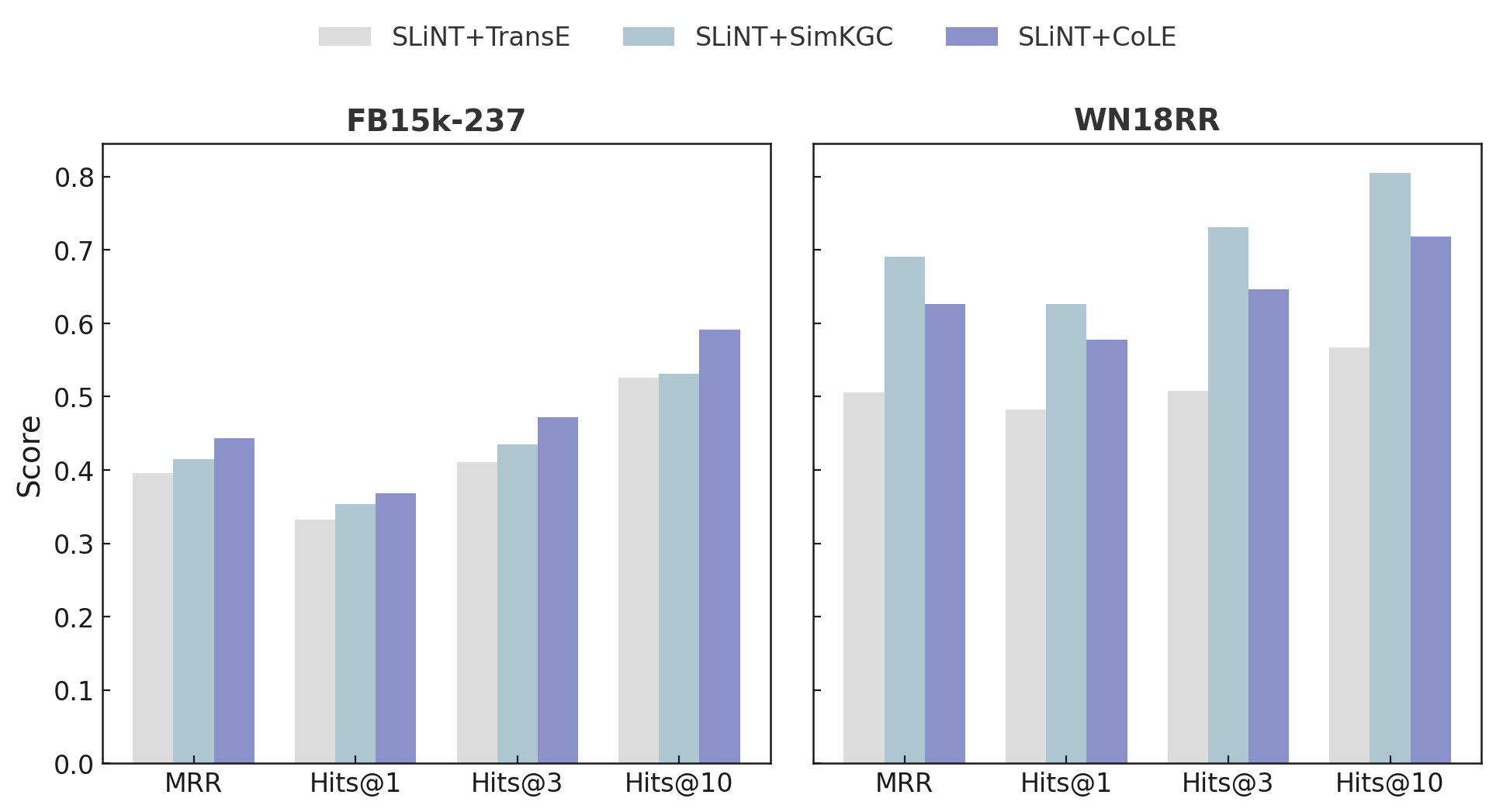}
    \caption{Comparing the performance of \textbf{SLiNT} with different structural encoders on FB15k-237 and WN18RR.}
    \label{fig:encoder_radar}
\end{figure}

\paragraph{Effect of Encoder Quality.}
SLiNT supports diverse structural encoders, and its performance can be further enhanced with more expressive ones. As shown in Table~\ref{tab:main_results_final} and Figure~\ref{fig:encoder_radar}, models using more powerful encoders such as CoLE consistently achieve higher performance. Notably, \textbf{SLiNT + CoLE} achieves the best MRR on FB15k-237, while \textbf{SLiNT + SimKGC} performs best on WN18RR, outperforming all baseline methods on their respective benchmarks.
 Even when paired with a simpler encoder like TransE, SLiNT surpasses several strong models, including LLaMA + TransE and SimKGC. These results demonstrate SLiNT's robustness to encoder quality and its ability to extract meaningful knowledge even from shallow embeddings.

\begin{figure}[t]
  \centering
  \includegraphics[width=1.0\linewidth]{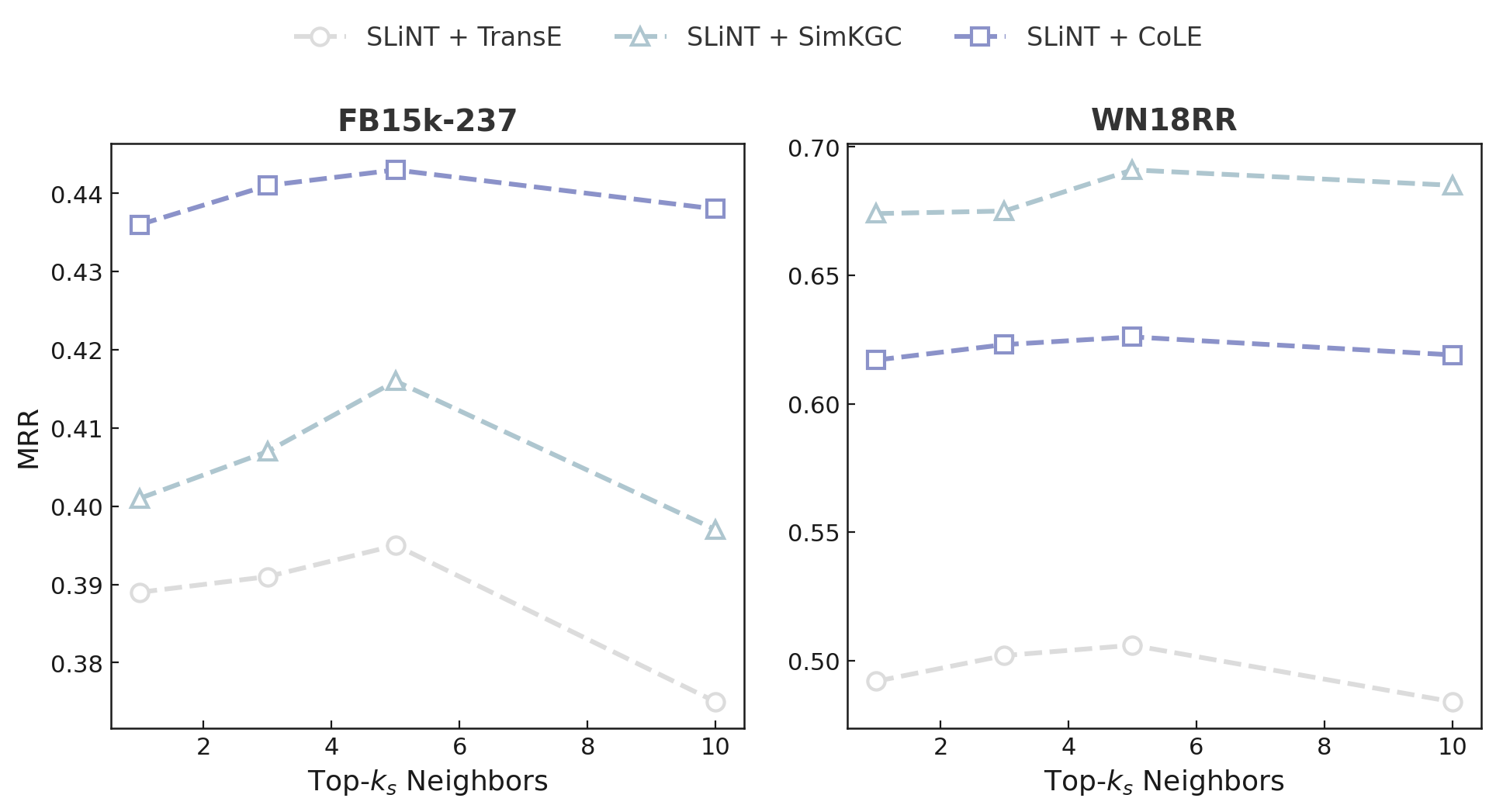}
  \caption{
  MRR comparison under varying Top-$k_s$ neighbor sizes for \textbf{SLiNT} with different structural encoders on FB15k-237 and WN18RR.
  }
  \label{fig:topk_encoder_twodatasets}
\end{figure}

\paragraph{Effect of Top-$k_s$ Neighbors.}
We evaluate how the number of structural neighbors ($k_s \in \{1, 3, 5, 10\}$) in SGNE affects performance across different encoders. As shown in Figure~\ref{fig:topk_encoder_twodatasets}, more expressive encoders like CoLE benefit steadily from increasing $k_s$, peaking at $k_s=5$. SimKGC shows similar trends but remains more stable. For weaker encoders like TransE, performance improves up to $k_s{=}5$ but drops at $k_s{=}10$ due to noisy neighbors. These findings suggest a trade-off: too few neighbors fail to capture the structure, while too many introduce noise, especially under less robust encoders. SGNE remains effective across these settings, with $k_s{=}5$ serving as a balanced default.

\paragraph{Effect of Contrastive Loss Weight.}
To assess the impact of contrastive supervision, we evaluate SLiNT under varying contrastive weights \(\lambda \in \{0.1, 0.3, 0.5, 0.7\}\) (Table~\ref{tab:lambda_ablation}). Results show that the optimal \(\lambda\) is dataset-specific: on FB15k-237, the best MRR and Hits@k scores are achieved at \(\lambda = 0.3\), while WN18RR reaches its peak at \(\lambda = 0.5\). This discrepancy likely reflects structural differences: WN18RR is more relation-regular with clearer decision boundaries, making it more receptive to contrastive supervision; In contrast, FB15k-237 contains more semantically overlapping relations, where overly strong contrastive signals may hinder generalization. Overall, these findings underscore the importance of balancing contrastive and generative objectives. Underweighting reduces the benefits of contrastive learning, while overweighting may destabilize training.
\begin{table}[t]
\centering
\renewcommand{\arraystretch}{1.1}
\setlength{\tabcolsep}{4.5pt}  
\scriptsize  
\begin{tabular}{c cccc|cccc}
\toprule
\multirow{2}{*}{\textbf{$\lambda$}} 
& \multicolumn{4}{c|}{\textbf{FB15k-237}} 
& \multicolumn{4}{c}{\textbf{WN18RR}} \\
& MRR & H@1 & H@3 & H@10 & MRR & H@1 & H@3 & H@10 \\
\midrule
0.1 & 0.438 & 0.354 & 0.462 & 0.580 & 0.617 & 0.571 & 0.639 & 0.709 \\
0.3 & \textbf{0.443} & \textbf{0.365} & \textbf{0.467} & \textbf{0.589} & 0.619& 0.574& 0.637& 0.708\\
0.5 & 0.437 & 0.352 & 0.459 & 0.582 & \textbf{0.626}&\textbf{0.578}& \textbf{0.646}&\textbf{0.718} \\
0.7 & 0.421 & 0.336 & 0.446 & 0.570 & 0.619 & 0.571 & 0.640 & 0.709 \\
\bottomrule
\end{tabular}
\vspace{-2mm}
\caption{
 Performance of \textbf{SLiNT + CoLE} under varying contrastive loss weights $\lambda$ on FB15k-237 and WN18RR.}
\label{tab:lambda_ablation}
\vspace{-2mm}
\end{table}

\section{Conclusion}
We present \textbf{SLiNT}, a structure-aware generative framework for knowledge graph completion that injects structure-derived evidence from KG embeddings into a frozen LLM backbone with lightweight LoRA-based adaptation. SLiNT integrates three complementary modules: SGNE for neighborhood enhancement, DHCL for dynamic contrastive supervision, and GDDI for token-level structure injection. Experiments on FB15k-237 and WN18RR demonstrate that SLiNT achieves superior or competitive performance, outperforming both embedding-based and generation-based baselines across multiple metrics, while maintaining robustness under structural sparsity and achieving efficient adaptation without full-model fine-tuning.

\section*{Limitations}
While \textbf{SLiNT} achieves strong performance on standard KGC benchmarks, it mainly leverages structure-derived signals from pretrained KG embeddings. This limits its applicability to scenarios requiring multimodal cues (e.g., images and temporal dynamics). Future work could extend SGNE to incorporate multimodal retrieval and design adaptive injection strategies for better generalization and efficiency in diverse real-world settings.

\section*{Ethical Consideration}
We use only publicly available datasets, which contain no private or sensitive information. No human subjects or annotation workers were involved. Our model incorporates open-source large language models (LLMs) for reasoning. While we do not fine-tune on sensitive content, we acknowledge potential risks of misuse or generation bias in downstream applications. 

\section*{Acknowledgments}
We would like to thank the reviewers and area chairs for their valuable feedback and constructive suggestions, which helped improve this work. 
This research was supported in part by the MIIT Project on Industrial Real-Time Database Based on Next-Generation Information Technology (TC210804D) and the CAS Project for Young Scientists in Basic Research (YSBR-040).



\bibliography{main}

\appendix
\section{Theoretical Justification}

\subsection{Complexity Analysis of Pseudo-Neighbor Retrieval}
\label{appendix:complexity}  
In our SGNE module, we retrieve the top-$k_s$ \textit{structural pseudo-neighbors} for each query or candidate entity by computing cosine similarity in the embedding space. Let the number of total entities be \textbf{N}, the embedding dimension be \textbf{d}, and the number of queries be \textbf{Q}. Then, for each query, the brute-force computation of cosine similarity has complexity:
\begin{equation}
\mathcal{O}(N \cdot d).
\end{equation}
Thus, the total retrieval cost becomes:
\begin{equation}
\mathcal{O}(Q \cdot N \cdot d).
\end{equation}

To reduce this cost in large-scale knowledge graphs, approximate nearest neighbor (ANN) methods such as FAISS can be used, reducing the retrieval complexity to:
\begin{equation}
\mathcal{O}(Q \cdot \log N \cdot d).
\end{equation}
This optimization enables scalable retrieval even for entity sets with millions of entries, as shown in \cite{DBLP:journals/tbd/JohnsonDJ21}. We further observe that the top-$k_s$ neighbors are precomputed and cached during training, making the cost negligible at inference time.

\subsection{Computational Cost Analysis}
\label{appendix:cost_analysis}

To assess the computational efficiency of \textbf{SLiNT}, we compare it against two representative baselines on the FB15k-237 dataset: (i) a vanilla \textbf{LLaMA} model without structural prompts or tuning, and (ii) \textbf{DIFT}, which incorporates structural embeddings via prompts but lacks our proposed modules (SGNE, DHCL, GDDI). We evaluate the training time per epoch and inference latency across three structural encoders (TransE, SimKGC, CoLE) on a single 64GB MetaX GPU. Table~\ref{tab:cost_analysis} summarizes the results.
\begin{table*}[h]
\centering
\small
\renewcommand{\arraystretch}{1.1}
\setlength{\tabcolsep}{6pt}
\begin{tabular}{l|ccc|c||ccc}
\toprule
\multirow{2}{*}{\textbf{Encoder}} 
& \multicolumn{3}{c|}{\textbf{Training Time (s/epoch)}} 
& \textbf{Overhead} 
& \multicolumn{3}{c}{\textbf{Inference Latency (ms/sample)}} \\
& \textbf{LLaMA} & \textbf{DIFT} & \textbf{SLiNT} & \textbf{vs. DIFT} 
& \textbf{LLaMA} & \textbf{DIFT} & \textbf{SLiNT} \\
\midrule
TransE  & 212.6 & 220.4 & 227.0 & +3.0\% & 9.3 & 9.7 & 10.1 \\
SimKGC  & 226.2 & 233.7 & 241.0 & +3.1\% & 9.7 & 10.1 & 10.5 \\
CoLE    & 231.5 & 239.3 & 246.6 & +3.1\% & 10.0 & 10.4 & 10.8 \\
\bottomrule
\end{tabular}
\caption{Training and inference cost of \textsc{SLiNT} vs. DIFT and LLaMA on FB15k-237.}
\label{tab:cost_analysis}
\end{table*}
\noindent
Although \textbf{SLiNT} integrates multiple modules—including pseudo-neighbor retrieval, attention-based fusion, contrastive learning, and token-level injection—it introduces only $\sim$3\% additional training time and less than 4\% inference latency overhead relative to the DIFT baseline. This efficiency is achieved through several design choices: (1) pseudo-neighbor retrieval is cached and reused during training; (2) the LLaMA backbone is kept frozen; and (3) LoRA adapters are lightweight and efficient. These results demonstrate that \textbf{SLiNT} is scalable and deployment-friendly without sacrificing computational efficiency.

\subsection{Robustness and Generalization of Contrastive Loss}
\label{appendix:contrastive}

To support fine-grained structural discrimination, our \textsc{DHCL} module optimizes a contrastive loss \(\mathcal{L}_{\text{CL}}\), as defined in Section~\ref{sec:dhcl}. This loss directly encodes structure-aware decision boundaries by comparing the distances between the query and interpolated hard positives/negatives in the structure space.

\paragraph{Stability via Lipschitz Continuity.}
Let \( Z_j = \| \tilde{\mathbf{q}} - \tilde{\mathbf{n}}_j \|_2 - \| \tilde{\mathbf{q}} - \tilde{\mathbf{p}}_j \|_2 \) denote the contrastive margin. Under unit-norm embeddings and interpolation sampling, we have \( Z_j \in [-2, 2] \), and the per-term loss \( \ell(Z_j) = -\log \sigma(Z_j) \) is 1-Lipschitz and bounded:
\[
\ell(Z_j) \in [\log(1 + e^{-2}),\ \log(1 + e^2)] \approx [0.13,\ 2.13].
\]
This boundedness ensures stable gradients and robustness, especially when interpolated negatives lie close to the decision boundary.

\paragraph{PAC-Bayes Generalization Bound.}
We adopt a PAC-Bayes analysis~\cite{DBLP:conf/colt/McAllester99, DBLP:conf/icml/SaunshiPAKK19} to study generalization under structure-sensitive contrastive training. Let the encoder \( f_\theta \) be parameterized by \( \theta \in \mathbb{R}^{d'} \), and assume a Gaussian prior \( P = \mathcal{N}(0, \sigma^2 I) \) and posterior \( Q = \mathcal{N}(\theta, \sigma^2 I) \). Then the expected risk satisfies:
\begin{equation}
\mathcal{R}(Q) \leq \hat{\mathcal{R}}(Q) 
+ \sqrt{\frac{1}{2N} \left(
\mathrm{KL}(Q \| P) + \log \frac{1}{\delta}
\right)},
\end{equation}

\noindent
with KL divergence given by:
\begin{equation}
\mathrm{KL}(Q \| P) = \frac{1}{2\sigma^2} \|\theta\|^2.
\end{equation}

Because the contrastive loss \(\mathcal{L}_{\text{CL}}\) is Lipschitz and bounded, it satisfies the assumptions of the PAC-Bayes framework. This result confirms that contrastive training under \textsc{DHCL} maintains stable generalization behavior even when interpolated negatives lie near structural decision boundaries.

\subsection{Modal Alignment Theory for Structure Injection}
\label{appendix:alignment}
Our GDDI  implements structure injection via \textit{token-level injection}, where the token embeddings of \([\text{QUERY}]\) and top-$k_r$ \([\text{ENTITY}]\) markers are replaced with structure-enhanced vectors $\tilde{\mathbf{q}}$ and $ \tilde{\mathbf{e}}_i$. These embeddings are injected into a frozen LLM, forming the input:
\begin{equation}
\mathbf{X}_{\text{input}} = [\text{CLS},\ \tilde{\mathbf{q}},\ \ldots,\ \tilde{\mathbf{e}}_{1},\ \ldots,\ \tilde{\mathbf{e}}_{k_r},\ \text{EOS}].
\end{equation}

We treat this process as a cross-modal alignment between graph-structured embeddings and language model token representations. Let the LLM be viewed as a conditional language model:
\begin{equation}
p_{\theta}(\mathbf{y} | \mathbf{X}).
\end{equation}
The structural embedding injection aims to preserve the semantic consistency between $\mathbf{X}_{\text{struct}}$ (the injected representation) and the output $\mathbf{y}$. Under the information bottleneck (IB) principle \cite{DBLP:conf/itw/TishbyZ15}, we define the learning objective as:
\begin{equation}
\max I(\mathbf{X}_{\text{struct}}; \mathbf{y}) - \beta I(\mathbf{X}_{\text{struct}}; \mathbf{Z}).
\end{equation}
where $\mathbf{Z}$ is the latent representation inside the LLM. This objective seeks a trade-off: inject structure such that it influences generation (high mutual info with $\mathbf{y}$), while not deviating excessively from LLM's internal representations.

In practice, we approximate this by replacing the token embeddings at designated slots with $\tilde{\mathbf{q}}$ and $ \tilde{\mathbf{e}}_i$, and minimizing the KL divergence between pre- and post-injection logits:
\begin{equation}
\mathcal{L}_{\text{align}} = \text{KL}(p_{\text{LM}}(\cdot|\mathbf{X}) \| p_{\text{LM}}(\cdot|\mathbf{X}_{\text{inject}}).
\end{equation}
This provides a differentiable surrogate for modal alignment. If the structure injection preserves or improves generation quality, we can conclude successful alignment.

\section{Dataset Overview }
\label{appendix:dataset-sparsity}
We conduct experiments on two widely used link prediction benchmarks: FB15k-237 and WN18RR. Table~\ref{tab:data_stats} summarizes the key dataset statistics.

\begin{itemize}
    \item \textbf{FB15k-237} is a refined subset of Freebase. It covers diverse entity types and relational patterns (e.g., 1-to-N, N-to-1). Redundant inverse edges are removed to avoid test leakage~\cite{DBLP:conf/emnlp/ToutanovaCPPCG15}.
    \item \textbf{WN18RR} is derived from WordNet, modeling lexical and hierarchical semantics such as hypernymy and derivation. Reversible relations are eliminated for robust evaluation~\cite{DBLP:conf/aaai/DettmersMS018}.
\end{itemize}

\begin{table}[h]
\centering
\small
\setlength{\tabcolsep}{5pt}
\renewcommand{\arraystretch}{1.05}
\begin{tabular}{lccccc}
\toprule
\textbf{Dataset} & \textbf{\#Ent.} & \textbf{\#Rel.} & \textbf{Train} & \textbf{Valid} & \textbf{Test} \\
\midrule
FB15k-237 & 14,541 & 237 & 272,115 & 17,535 & 20,466 \\
WN18RR    & 40,943 & 11  & 86,835  & 3,034  & 3,134 \\
\bottomrule
\end{tabular}
\caption{Basic statistics of benchmark datasets.}
\label{tab:data_stats}
\end{table}

\section{Training and Implementation Details}
\label{appendix:training}

We provide detailed configuration settings for all components of \textbf{SLiNT}.

\paragraph{Hardware and Framework.}
All experiments are conducted on 8×64GB MetaX GPUs, a domestic CUDA-compatible accelerator with performance comparable to NVIDIA A100s. Our implementation is based on PyTorch with automatic mixed-precision (AMP) training for efficiency.

\paragraph{Backbone Model.}
We use the frozen LLaMA-2-7B model from HuggingFace\footnote{\url{https://huggingface.co/meta-llama/Llama-2-7b-chat-hf}} as the base language model. Structure-aware features are injected via prompt and token-level replacements without updating LLM parameters. LoRA is used for efficient adaptation, with rank $r = 128$, scaling factor $\alpha = 64$, and dropout rate of 0.1.

\paragraph{KG Embeddings.}
We experiment with three pretrained KG encoders: TransE, SimKGC, and CoLE. Each query is used to retrieve a top-$m$ candidate list from a pretrained KG embedding model, with $m=20$. These embeddings provide the structural foundation for neighborhood enhancement and contrastive supervision.

\paragraph{SGNE Settings.}
In the SGNE module, we retrieve top-$k_s = 5$ pseudo-neighbors for each query or candidate entity using cosine similarity in the KG embedding space. The query and its neighbors are fused with multi-head attention, and the outputs are cached for efficiency.

\paragraph{DHCL Settings.}
For contrastive training, we sample $N = 50$ candidate entities per query. The contrastive loss is computed over $k_c = 10$ hard positives and negatives selected based on pseudo-neighbor overlap. The loss is weighted by a confusion-aware scoring function. The contrastive loss coefficient is $\lambda = 0.5$.

\paragraph{GDDI Settings.}
In the GDDI module, we inject $k_r = 1$ structure-enhanced entity token into each input sequence. The enhanced embeddings replace the placeholders for \texttt{[QUERY]} and \texttt{[ENTITY]} tokens. Injection is performed at both prompt-level (text) and token-level (embedding) positions.

\paragraph{Optimization and Training.}
All models are trained using the Adam optimizer with a learning rate of $2 \times 10^{-5}$ and a batch size of 64. We train for 3 epochs with early stopping based on validation MRR. Random seeds are fixed for reproducibility, and all results are averaged over three runs.

\section{Case Study}
\label{appendix:case}

To demonstrate how SLiNT leverages pseudo-neighbor injection and contrastive learning, we present three representative cases from WN18RR and FB15k-237. Each case includes a query, candidates, SGNE-retrieved pseudo-neighbors, and model predictions. Tables~\ref{tab:case_tail}–\ref{tab:case_fail} are presented alongside their respective discussions.

\vspace{1mm}
\subsection{Case 1: Disambiguating Musical Components}

\begin{table}[H]
\centering
\small
\renewcommand{\arraystretch}{1.1}
\setlength{\tabcolsep}{6pt}
\begin{tabularx}{\linewidth}{@{}p{2.8cm}X@{}}
\toprule
\textbf{Query} & (instrument, has\_part, ?) \\
\textbf{Candidates} & \{bow, string, keyboard, bridge\} \\
\textbf{Ground Truth} & string \\
\textbf{Pseudo-Neighbors (Top-5)} & violin, cello, harp, guitar, banjo \\
\midrule
\textbf{LLaMA + CoLE} & keyboard (Top-1) \\
\textbf{DIFT + CoLE} & keyboard (Top-1) \\
\textbf{SLiNT + CoLE} & \textbf{string} (Top-1) \\
\bottomrule
\end{tabularx}
\vspace{-1mm}
\caption{Case 1: SLiNT correctly predicts \texttt{string} by leveraging structurally similar instruments.}
\label{tab:case_tail}
\end{table}

\noindent\textbf{Analysis.} LLaMA and DIFT choose \texttt{keyboard}, a plausible but structurally irrelevant part of KG. SLiNT identifies \texttt{string} by leveraging pseudo-neighbors such as \texttt{violin} and \texttt{cello}, where \texttt{string} is a shared component.

\vspace{2mm}
\subsection{Case 2: Differentiating Geopolitical Containment}

\begin{table}[H]
\centering
\small
\renewcommand{\arraystretch}{1.1}
\setlength{\tabcolsep}{6pt}
\begin{tabularx}{\linewidth}{@{}p{2.8cm}X@{}}
\toprule
\textbf{Query} & (?, location\_contains, mountain) \\
\textbf{Candidates} & \{Nepal, Asia, Everest, Tibet\} \\
\textbf{Ground Truth} & Nepal \\
\textbf{Pseudo-Neighbors (Top-5)} & Himalaya, Kathmandu, Pokhara, Lumbini, Mustang \\
\midrule
\textbf{LLaMA + CoLE} & Asia (Top-1) \\
\textbf{DIFT + CoLE} & \textbf{Nepal} (Top-1) \\
\textbf{SLiNT + CoLE} & \textbf{Nepal} (Top-1) \\
\bottomrule
\end{tabularx}
\vspace{-1mm}
\caption{Case 2: SLiNT correctly identifies \texttt{Nepal} by grounding in local structural cues.}
\label{tab:case_geo}
\end{table}

\noindent\textbf{Analysis.} Although all candidates are semantically relevant to \texttt{mountain}, SLiNT uses structure-guided cues, e.g., \texttt{Himalaya} and \texttt{Kathmandu} to localize the correct geopolitical scope.

\vspace{2mm}
\subsection{Case 3: Failure Case on Long-tail Relation}

\begin{table}[H]
\centering
\small
\renewcommand{\arraystretch}{1.1}
\setlength{\tabcolsep}{6pt}
\begin{tabularx}{\linewidth}{@{}p{2.8cm}X@{}}
\toprule
\textbf{Query} & (person, known\_for, ?) \\
\textbf{Candidates} & \{acting, painting, novel, photography\} \\
\textbf{Ground Truth} & painting \\
\textbf{Pseudo-Neighbors (Top-5)} & artist, sculptor, painter, curator, illustrator \\
\midrule
\textbf{LLaMA + CoLE} & acting (Top-1) \\
\textbf{DIFT + CoLE} & acting (Top-1) \\
\textbf{SLiNT + CoLE} & novel (Top-1) \\
\bottomrule
\end{tabularx}
\vspace{-1mm}
\caption{Case 3: SLiNT fails to predict \texttt{painting}, despite partial structural grounding.}
\label{tab:case_fail}
\end{table}

\noindent\textbf{Analysis.} All models fail to predict \texttt{painting}, revealing challenges in handling long-tail relations. SLiNT ranks \texttt{novel} highest, likely influenced by relevant creative-profession neighbors, yet fails to fully disambiguate the semantic role of the candidate. This failure highlights the limitations of structure-based grounding in the absence of sufficient semantic alignment.

\end{document}